# Image denoising by Super Neurons: *Why go deep?*

Junaid Malik[a], Serkan Kiranyaz[b], Moncef Gabbouj[a]

[a]*Tampere University, Tampere, Finland*
[b]*Qatar University, Doha, Qatar*

**Abstract**

Classical image denoising methods utilize the non-local self-similarity principle to effectively recover image content from noisy images. Current state-of-the-art methods use deep convolutional neural networks (CNNs) to effectively learn the mapping from noisy to clean images. Deep denoising CNNs manifest a high learning capacity and integrate non-local information owing to the large receptive field yielded by numerous cascade of hidden layers. However, deep networks are also computationally complex and require large data for training. To address these issues, this study draws the focus on the Self-organized Operational Neural Networks (Self-ONNs) empowered by a novel neuron model that can achieve a similar or better denoising performance with a compact and shallow model. Recently, the concept of super-neurons has been introduced which augment the non-linear transformations of generative neurons by utilizing non-localized kernel locations for an enhanced receptive field size. This is the key accomplishment which renders the need for a deep network configuration. As the integration of non-local information is known to benefit denoising, in this work we investigate the use of super neurons for both synthetic and real-world image denoising. We also discuss the practical issues in implementing the super neuron model on GPUs and propose a trade-off between the heterogeneity of non-localized operations and computational complexity. Our results demonstrate that with the same width and depth, Self-ONNs with super neurons provide a significant boost of denoising performance over the networks with generative and convolutional neurons for both denoising tasks. Moreover, results demonstrate that Self-ONNs with super neurons can achieve a competitive and superior synthetic denoising performances than well-known deep CNN denoisers for synthetic and real-world denoising, respectively.

## 1. Introduction

Image denoising is a widely studied low-level inverse imaging task that concerns removing noise from digital images. One of the key principles that drive traditionally successful image denoising methods is the integration of non-local information towards recovering a particular area of image [1], [2]. These methods exploit the non-local self-similarity (NSS) principle prior which states that natural images tend to have a repetition of textures and patterns spread around different locations in the image. When denoising a particular patch of a noisy image, methods exploiting NSS search for similar non-local regions and use the information from those regions to recover the content of the noisy patch more effectively. In recent times, convolutional neural networks (CNNs) have started to dominate the field of denoising by directly learning to denoise images using a set of noisy-clean image pairs. CNNs for image denoising are mostly deep, i.e., they contain many layers of convolutional neurons, each of which applies the sliding window-based spatial convolution operation. The success of deep CNNs for image denoising can be attributed to two key properties: i) the non-linear modeling capacity yielded by repeating blocks of convolution and a non-linear activation function, and ii) enhanced receptive field attained by stacking multiple layers of convolutional neurons. While not directly utilizing NSS like the previous methods, CNNs nevertheless exploit the non-local information by the virtue of their enhanced receptive field.

The receptive field of a CNN is directly correlated with the depth of the network [3]. However, constraints on the computational complexity, especially considering that denoising networks require a full image-to-image mapping, put practical limits on the number of layers. Various structural enhancements to the standard CNN-based denoising strategy have been explored in recent denoising works. In [4], the concept of dilated convolutions is utilized which increases receptive field without introducing additional trainable parameters. In [5], down-sampled sub-images are utilized to increase the effective receptive field. In [6], a recursive CNN structure is proposed and in [7], repeated up-down sampling blocks inspired by [8] were used. While sampling feature maps sparsely as in dilation is known to cause artifacts in image and may not always be beneficial [5], down-sampling feature maps is not ideal for full image-to-image mapping, because it necessitates the learning of the reverse operation for up-sampling. Moreover, all the fore-mentioned works still utilize deep networks, with as many as 80-layers as in [6]. The direct impact of enhancing the receptive field on image denoising in such deep network configurations is not conclusive as the superior denoising performance can be attributed to the increased model capacity that results from a deep configuration. Finally, none of the works propose a unified scheme for enhancing the model capacity and the receptive field simultaneously. Recently, interesting developments have been made in this regard in the pursuit of better neuron models.

In [9], the generative neuron model was introduced as an alternative to convolutional neurons where the kernel transformations (i.e., the nodal operators) are generated during training based on the Maclaurin series approximations. The generative neurons powered the so-called Self-Organized Operational Neural Networks (Self-ONNs) which significantly surpassed the equivalent, and sometimes even more complex/deeper CNNs over a variety of tasks [10] and especially in denoising [11]. This was attributed mainly to the increased modeling capacity induced by the embedded non-linearities optimized for the learning task in hand. Most recently, building upon the generative neuron model, the concept of "super (generative) neurons" was introduced in [12]. Super neurons propose non-localized kernels over the generative neuron model so that they can gather information from a larger area in the previous layer feature maps while keeping the kernel size as is. As the first model, kernel locations are randomized and the each kernel transformation (nodal operator) is optimized during the BP training for that particular location. It is sometimes more desirable "to learn" the optimal kernel locations *during* the training process along with the generated (nodal) operators. Therefore, as the second model, both kernel location and its (nodal) operation are optimized simultaneously. As a proof-of-concept, the study in [12] demonstrated that when trained under fair conditions over a scarce train data, compact Self-ONNs with super neurons were shown to surpass the Self-ONNs with generative neurons significantly over a variety of regression tasks.

In this study, the primary motivation stems from the fact that non-localized kernels effectively enable the integration of non-local information, which is of particular importance for image denoising applications. Therefore, we investigate the practical implementation of super neuron-based models for image denoising and perform extensive comparisons over various severity levels of synthetic noise as well as real-world blind denoising. Moreover, we analyze the impact of the enhanced receptive fields by visualizing the effective receptive fields of the trained models. The main contributions of this work are listed below:

1. We evaluate the super-neuron model for synthetic and real-world image denoising and perform comparisons with equivalent Self-ONNs, CNNs, and a deep CNN-based denoiser, the DnCNN [13].
2. We further discuss other practical issues regarding a GPU-based implementation of super-neuron-based models and introduce a relaxation to balance the enhancement of receptive fields and computational efficiency.
3. We analyze the impact of increasing the receptive field by visualizing the effective receptive fields for each of the trained networks.

The remainder of this paper is organized as follows: Section 2 provides a brief literature review of the relevance of receptive fields for image denoising methods and the operational neuron model which has evolved to produce the recent super neuron model. In Section 3, the formulation behind image denoising with super neurons is provided along with the practical considerations for implementation. Section 4 provides technical details concerning the experimental settings while Section 5, presents all the denoising results with both quantitative and qualitative comparative evaluations. Finally, Section 6 discusses the results, performance comparisons and possible future directions. Section 7 concludes the paper.

## 2. Related Work

### 2.1. Image Denoising and Receptive Fields

The presence of noise deters the correlation naturally present between the pixels inside local neighborhoods in images. This is one of the main reasons why denoising methods employing non-local information have achieved significant success over the years [1], [2]. Particularly, successful techniques are built upon the non-local self-similarity (NSS) prior, which assumes that natural images generally have repetitive features at different locations in the image. By exploiting data from these similar non-local areas of the noisy image, the original image content can be recovered more effectively. Although convolution is a local operation, non-local information is encoded in deep CNN-based denoisers in the form of a larger receptive field obtained by the stacking of convolution layers. Increasing the depth of the model is the most straightforward way of increasing the receptive field [3]. Still, this comes at an expense of additional parameters and requires rich training resources. Another approach for expanding the receptive field is by pooling or down-sampling the feature maps. However, for image denoising, the input and output resolution must be the same and any pooling (down-sampling) operation must be subsequently reversed using up-sampling which can be taxing for the network to learn properly and makes the output prone to artifacts. Several CNN-based denoising studies utilize variants of convolution operation to yield a larger effective receptive field while minimizing the impact on efficiency and modeling capacity. In [4], dilated convolutions are used to increase the receptive field size of a 10-layer network and achieve denoising performance at par with a 17-layer network. In [6], a recursive multi-level CNN is used to increase the receptive field size and achieve state-of-the-art denoising performance with an 80-layer network. In [5], denoising is applied on down-sampled sub-images to increase the size of the receptive field and a sub-pixel convolution layer is used to reverse the down-sampling.

### 2.2. Operational Neural Networks

To address the limited diversity achieved by the sole linear neuron model of CNNs, operational neural networks (ONNs) were proposed in [9]. The core principle behind ONNs is to enable a heterogenous network model with any (linear or non-linear) operators. Specifically, given a 2D input $x$ and the weights $w$, the output $y$ of the convolution and operational neurons are expressed in (1) and (2) respectively.

$$\textbf{\textit{Convolutional Neuron:}} \; y(m,n) = \sum_{r=0}^{K-1}\sum_{t=0}^{K-1} w(r,t)x(m+r, n+t) \tag{1}$$

$$\textbf{\textit{Operational Neuron:}} \quad y(m,n) = P\left(\psi(w(r,t), x(m+r, n+t))\right)_{(r,t)=(0,0)}^{(K-1,K-1)} \quad (2)$$

where $P$ and $\psi$ are pool and nodal operators, respectively. Provided a suitable operator set library and a policy to assign an appropriate operator set to each neuron [14], ONNs were shown to outperform their CNN counterparts on different proof-of-concept regression applications. However, the extension of ONNs to deeper and more complex models is hindered by the need of defining a library of suitable operator sets for each application and assigning one to each neuron/layer of the network, which necessitates additional training runs. As a workaround, self-organized ONNs based on the *generative* neuron model (Self-ONNs) [10] were proposed which utilize a Maclaurin polynomial of degree Q to generate any non-linear kernel (nodal) operation function.

$$\textbf{\textit{Generative Neuron:}} \quad y(m,n) = P\left(\sum_{q=1}^{Q} w^{(q)}(r, t, \boldsymbol{q})(x(m+r, n+t))^q\right)_{(r,t)=(0,0)}^{(K-1,K-1)} \quad (3)$$

Despite requiring additional trainable parameters as compared to an equivalent convolutional neuron, Self-ONNs were shown to surpass CNNs in several tasks, especially for image denoising even with equivalent computational requirements [11].

*2.2.1.1. Super Neurons*

While generative neurons of Self-ONNs are superset of the convolutional neurons and enable non-linear kernel transformations when $Q>1$, it is nevertheless key to note that the receptive fields of both Self-ONNs and CNNs are still bounded by the depth of the network. For compact networks to compete effectively with their deeper counterparts, introducing non-linear transformations alone might not suffice as the receptive fields remain limited. To remedy this, the concept of a super neurons [12] has recently emerged which builds upon the foundations of a generative neuron and extends it by enabling *non-localized* kernel transformations as expressed below:

$$\textbf{\textit{Super Neuron:}} \quad y(m,n) = P\left(\sum_{q=1}^{Q} w^{(q)}(r, t, q)(x(m+r+\boldsymbol{\alpha}, n+t+\boldsymbol{\beta}))^q\right)_{(r,t)=(0,0)}^{(K-1,K-1)} \quad (4)$$

By shifting the center of the kernel by $\alpha$ and $\beta$ in the $x$ and $y$ direction respectively, super neurons effectively operate on a larger area of input feature maps, thus increasing the receptive field without needing additional parameters.

## 3. Methodology

*3.1. Image Denoising with Super Neurons*

A standard modeling notation for the image denoising problem is to formulate the noisy image as:

$$x_{noisy} = x + n \quad (5)$$

where $x_{noisy}$ is the noisy image, $x$ is the clean image and $n$ is the noise. The most common assumption is for $n$ to be data-independent Additive White Gaussian Noise (AWGN) with a certain standard deviation of $\sigma$. In non-blind CNN-based denoising methods, the clean image is estimated as:

$$x_{estimated} = F(x_{noisy}, \sigma, \theta) \quad (6)$$

where $x_{estimated}$ is the denoised image, $F$ represents the model, $\sigma$ is the standard deviation of the noise and $\theta$ represents the trainable parameters of the model. The network is trained on a set noisy-clean image pairs to obtain the optimal set of weights $\hat{\theta}$ which minimize a loss function $loss\left(x, F(x_{noisy}, \sigma, \theta)\right)$ across the entire training set. In a simple feed-forward setting, the model $F$ is composed of $L$ layers and each layer $l$ is composed of $N_l$ neurons. The output of a neuron $k$ in layer $l$ is represented as follows:

$$y_l^k = \sum_{j=1}^{N_{l-1}} \phi_l^k(y_{l-1}^j * w_{l-1}^{jk}) \quad (7)$$

where $\phi_l^k$ is the pointwise non-linear activation function. For the super-neurons, the output is further parameterized by the order of the Maclaurin polynomial $Q$ as well as the pairs of kernel shifts $(\alpha_l^{jk}, \beta_l^{jk})$, and computed as shown in (4). In practice, the shifts $\alpha_l^{jk}$ and $\beta_l^{jk}$ can be applied to the previous layer feature maps instead and bounded by a suitable range of $\pm\gamma_x$ and $\pm\gamma_y$ respectively. The shifted feature maps are generated using bilinear interpolation as shifts are allowed to take any real value. For detailed pixel-wise forward and back-propagation through super neurons, the reader is referred to [12].

*3.2. Computational Aspects for Parallelized Super Neuron Implementation*

Earlier, in [15], a parallelized GPU implementation of a particular case of *generative* neuron model (when the pooling function is set to summation) was achieved by concatenating higher powers of input tensors to create an expanded tensor and utilizing a single convolution operation to obtain the layer output (for the detailed formulation, the reader is referred to [15]). This was possible because the previous layer exponentiated feature maps were reused by all neurons in the current layer. This property of generative neurons, despite the additional training parameters, makes them computationally effective as compared to operational neurons because it lends itself to an parallelized implementation on GPUs. However, the formulation presented in [15] is not directly applicable for the case of super neurons because the output of a previous layer neuron, $y_{l-1}^j$, is shifted uniquely by $(\alpha_l^{jk}, \beta_l^{jk})$ for each of the $k$ connections to the neurons in the next layer. Leveraging convolution operation as in the generative neuron case would require an expanded tensor of shape $B \times (N_{l-1} \times N_l \times Q) \times H \times W$ for each layer-to-layer connection, which requires substantial GPU memory and renders the training and inference process prohibitively expensive. As a workaround to this, we propose a trade-off between kernel shifts and computational efficiency by sharing the shifts among the neurons of the current layer. Mathematically, this would be expressed as, $\alpha_l^{jk} = \alpha_l^j$ and $\beta_l^{jk} = \beta_l^j$. In this way, a layer with such *partial* super neurons can be implemented as a cascade of batch-wise shifting operation with $(\alpha_l^j, \beta_l^j)$ followed by a Self-ONN layer. This is visualized in Figure 1.

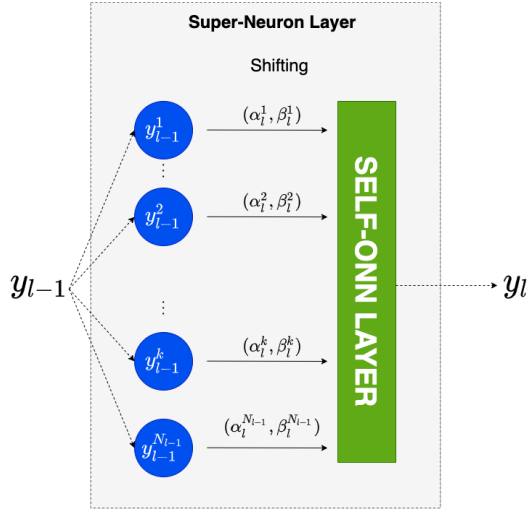

Figure 1. The proposed partial super-neuron layer leverages the Self-ONN layer by sharing shifted feature maps between neurons.

The bilinear interpolation operation is implemented using the grid sampling functionality offered in Pytorch [16]. The required shifts are represented as an affine transformation matrix that, coupled with the information about the feature map size, is utilized to create a sampling grid that encodes the shifted coordinates. Finally, the data is sampled from the feature maps using bilinear interpolation. All these operations are implemented on GPU and are automatically differentiable.

## 4. Experiments

### 4.1. Noise models

For training and comprehensive evaluation of the networks used in this study, we employ both synthetic and real noise models. For the former, Additive White Gaussian Noise (AWGN) with standard deviation $\sigma$ is used to synthetically corrupt the clean images. The noisy images are also clipped between the standard range of 0 and 1. Unlike this synthetic noise model, the absence of a clean image in the case of real-world noisy images requires a certain ground-truth estimation procedure to produce noisy-clean image pairs. For this purpose, we use noisy and clean image pairs from the benchmark SIDD dataset [17].

### 4.2. Network Architecture

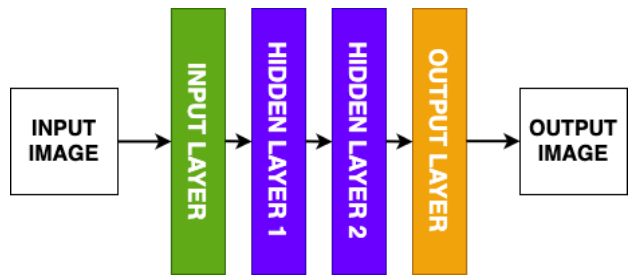

Figure 2. The architecture of the compact networks used in this study.

To highlight the impact of enhanced receptive fields, we use compact networks with 2 hidden layers and utilize all three neuron models: convolutional, generative, and super neurons. For convolutional neurons, a 17-layer architecture is also used to obtain a comparison with deep networks. All networks used in the study were trained from scratch using the same training data and hyperparameters. For Self-ONNs with generative neurons, the order of Maclaurin polynomial, $Q$, is within the range [3,5,7]. For super neurons, the value of $Q$ is 3, and the maximum absolute limit of shifts $\gamma_x$ and $\gamma_y$ are set as 5. In this study, we employ random bias mode of super neurons where all spatial bias pairs, $(\alpha_l^j, \beta_l^j)$, in a Super-ONN are sampled uniformly within the interval $(-5, +5)$ for the hidden layers and set to zero for the input and output layers of the network. As a simplified terminology, from now on we will call the Self-ONNs with the super neurons as "Super-ONNs". The architectural characteristics of the models are presented in detail in Table 1.

Table 1. Architectural details of the networks used in this study.

| Network | Layers | Width ($W$) | Neuron Model | $Q$ | $(\gamma_x, \gamma_y)$ |
|---|---|---|---|---|---|
| CNN | 4 | 128 | Convolutional | 1 | - |
| Self-ONN-$Q$-$W$ | 4 | 64 | Generative | [3,5,7] | - |
| Super-ONN-$Q$-$W$ | 4 | 64 | Super | [3,5,7] | $\mathcal{U}(-5,+5)$ |
| DnCNN | 17 | 64 | Convolutional | 1 | - |

*4.3. Training Settings*

The training data for synthetic denoising is the same as that used in [18] consisting of approximately 200k patches of size 40x40 cropped from the BSD400 dataset. High-resolution images from Kodak, McMaster and BSD68 datasets are used for testing. In real-world denoising experiments, we utilize the SIDD Medium training dataset which consists of 320 high-resolution images. We use the same cropping strategy as adopted in [15] to extract 160k training patches. For testing, the SIDD validation dataset is used which consists of 1280 noisy clean image pairs. In all the experiments, the training to validation ratio is set to 9:1.

For gradient descent, we use the ADAM optimizer with the maximum learning rate set to $10^{-3}$. All the networks were trained for 100 epochs and the model state which maximized the validation set performance was chosen for evaluation. Model architectures were defined using FastONN [19] library and Pytorch library [20]. All experiments were performed either on an NVIDIA Tesla V100 or an NVIDIA TITAN RTX GPU.

## 5. Results

In this section, the results of the two denoising problems tackled will be presented along with an interpretation of the results. All comparisons are based on the Peak Signal to Noise Ratio (PSNR) of the competing models.

*5.1. Synthetic Denoising*

The quantitative results for synthetic image denoising in terms of average PSNR levels are presented in Table 2 and the visual results are shown in Figure 3. Except BM3D [ref] and DnCNN [ref], all networks have the 2-hidden layer architecture as shown in Figure 2. For simplicity, we called a Self-ONN with super neurons as "Super-ONN". As reported in [18], Self-ONNs with generative neurons can surpass both CNNs and BM3D in all datasets and noise levels. Here we analyze how the super neuron model fares when compared to the generative model. Averaged across the three evaluation datasets, the best performing Super-ONN outperforms its Self-ONN counterpart by 0.34 dB for $\sigma = 30$, 0.72dB for $\sigma = 60$ and 0.93 dB for $\sigma = 90$ noise. If the networks with the same $Q$ values are compared, super neurons consistently achieve an improvement over the Self-ONN with the same $Q$ value, and the improvement increases with the severity of the noise. Furthermore, the Super-ONN model with the least model complexity (Super-ONN-3-64) regularly surpasses the performance of the most complex Self-ONN (Self-ONN-7-64) across all noise levels and datasets. Obviously, the performance gap between CNN and Self-ONNs further widens with the super neurons regardless from the $Q$ setting.

Table 2. Average PSNR levels of AWGN denoising for BM3D [ref], DnCNN with 17 layers [ref], and denoising networks with 2 hidden layers (CNN, Self-ONN and Super-ONN) over three benchmark datasets and three noise levels.

| | Param. No (k) | KODAK | | | McMaster | | | CBSD68 | | |
|---|---|---|---|---|---|---|---|---|---|---|
| | | σ=30 | σ=60 | σ=90 | σ=30 | σ=60 | σ=90 | σ=30 | σ=60 | σ=90 |
| BM3D [21] | | 28,58 | 25,05 | 22,44 | 29,30 | 24,76 | 21,59 | 27,17 | 23,65 | 21,27 |
| CNN-128 | 297,60 | 28,47 | 25,08 | 23,11 | 29,28 | 25,73 | 23,55 | 27,47 | 24,28 | 22,43 |
| Self-ONN-3-64 | 224,83 | 28,54 | 25,09 | 23,12 | 29,39 | 25,77 | 23,66 | 27,55 | 24,31 | 22,47 |
| Self-ONN-5-64 | 374,59 | 28,56 | 25,10 | 23,12 | 29,40 | 25,81 | 23,65 | 27,55 | 24,34 | 22,46 |
| Self-ONN-7-64 | 524,35 | 28,54 | 25,12 | 23,12 | 29,40 | 25,77 | 23,64 | 27,56 | 24,33 | 22,46 |
| Super-ONN-3-64 | 224,83 | 28,84 | 26,18 | 24,51 | 29,58 | 26,13 | 24,13 | 27,90 | 24,93 | 23,38 |
| Super-ONN-5-64 | 374,59 | 29,06 | 26,21 | 24,53 | 29,51 | 26,29 | 23,90 | 27,76 | 24,93 | 23,14 |
| Super-ONN-7-64 | 524,35 | 29,05 | 26,20 | 24,42 | 29,45 | 26,08 | 23,73 | 27,72 | 24,86 | 23,20 |
| DnCNN [13] | 556,03 | 29,43 | 26,52 | 24,91 | 30,27 | 26,97 | 25,08 | 28,24 | 25,34 | 23,78 |

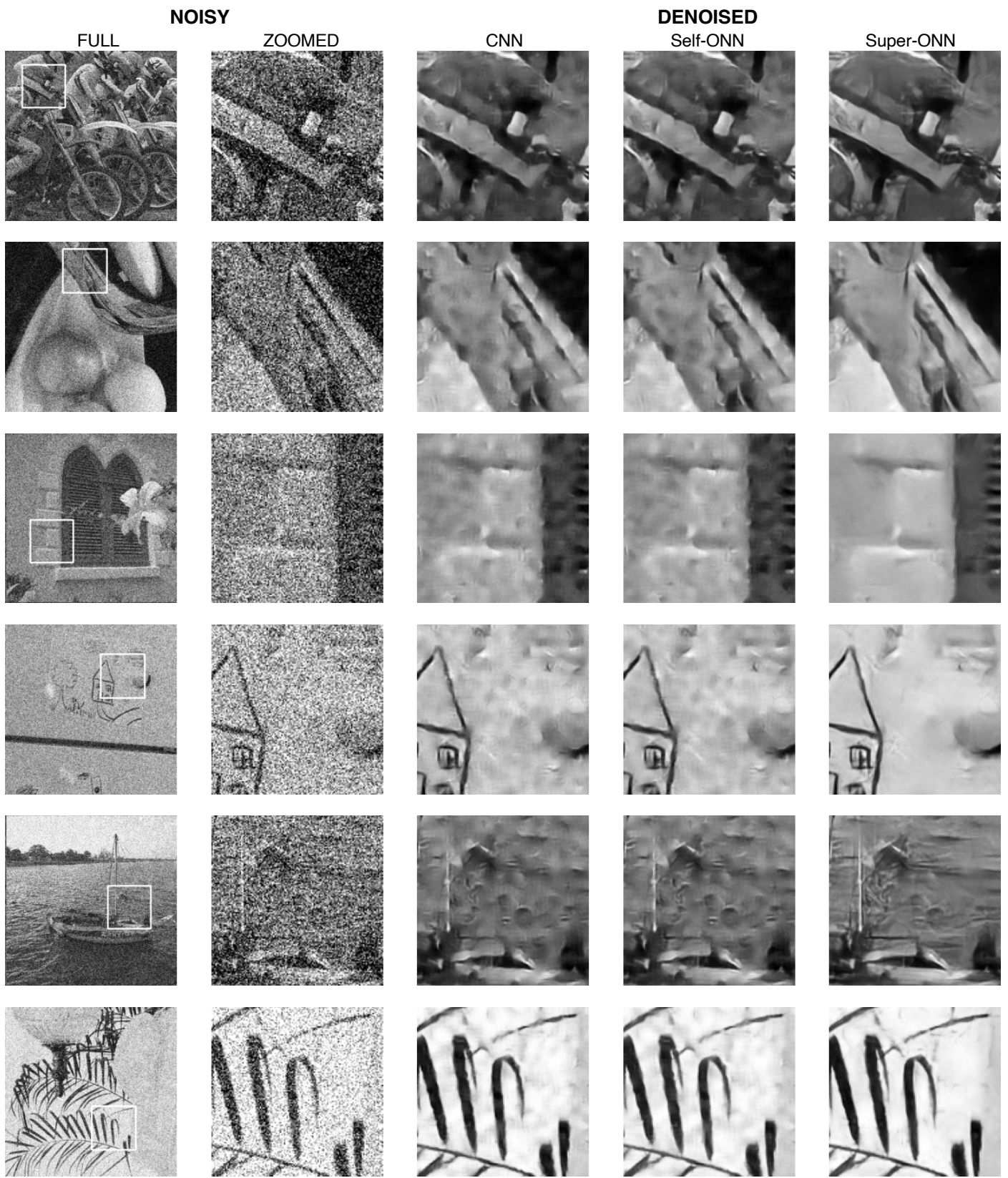

Figure 3. Some AWGN corrupted images (left), their zoomed sections (2nd column) and the corresponding outputs of the CNN (3rd column), Self-ONN (4th column) and Super-ONN (right) from the test partition.

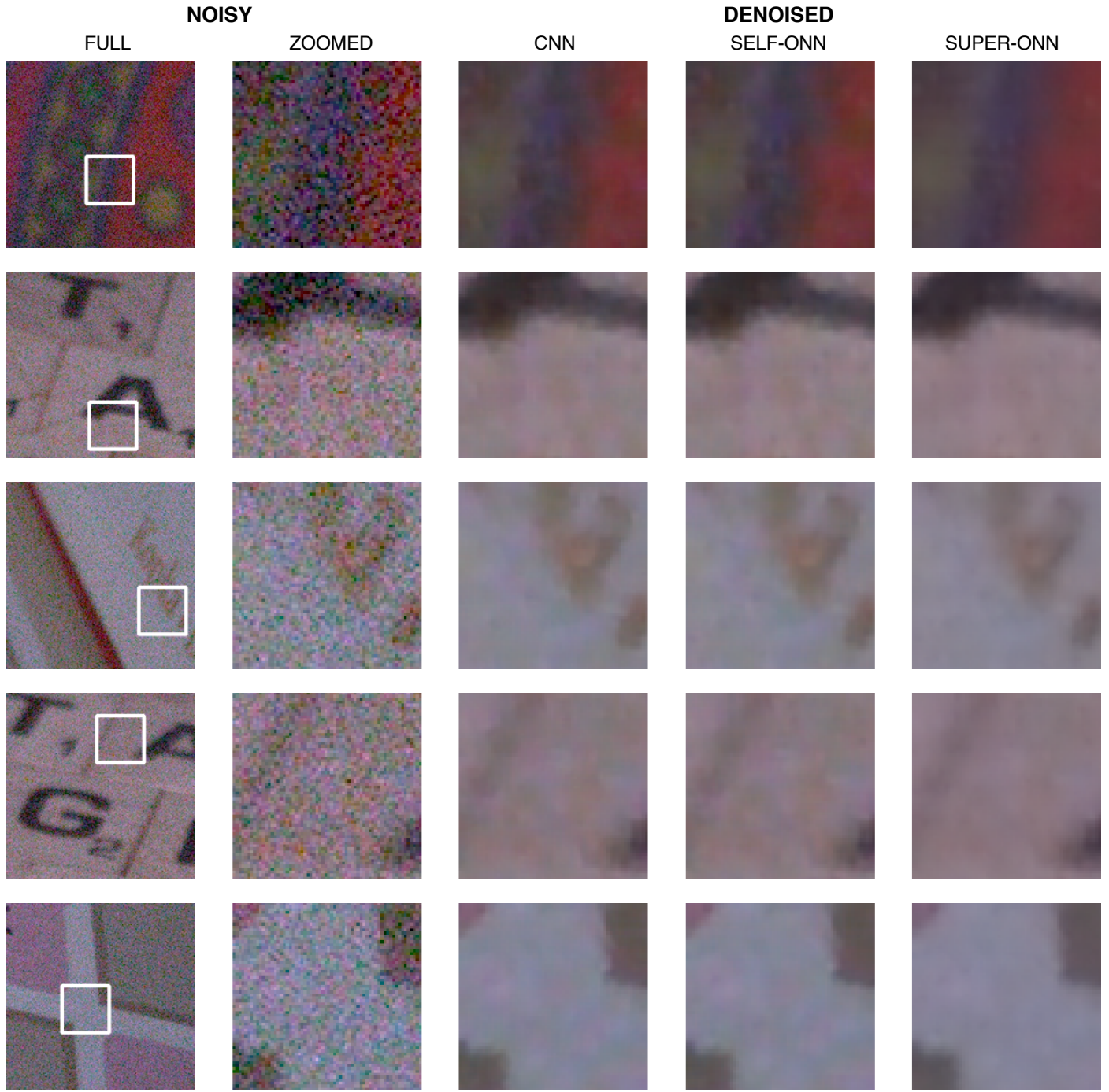

Figure 4. Some images with real-world noise (left), their zoomed sections (2nd column) and the corresponding outputs of the CNN (3rd column), Self-ONN (4th column) and Super-ONN (right) from the test partition of SIDD dataset.

Lastly, the 17-layer deep CNN provides an improvement of 0.46 dB for both $\sigma = 30$ and $\sigma = 60$ and 0.57dB for $\sigma = 90$ noise over the best super neuron model, averaged across the three datasets. Qualitatively, the superior denoising performance of Self-ONNs with super-neurons is visible in all output images shown in Figure 3. In particular, Super-ONNs not only show a sharper edge and texture details but also recover the smooth regions much better with fewer noticeable artifacts.

*5.2. Real World Denoising*

The quantitative results for the real-world denoising problem in terms of average PSNR levels are presented in **Error! Not a valid bookmark self-reference.** and the visual results on the SIDD Validation dataset are shown in Figure 4. The best Super-ONN outperforms the best performing Self-ONN by a margin of 1.76dB (4.95%). Moreover, the Super-ONNs with 2 hidden-layers also outperforms the 8-layer Dn-SelfONN of [15] by a margin of 1.04 dB. For the comparison with generative neurons of the same $Q$ value, Super-ONNs achieve improvements of 1.81dB, 1.49dB, and 1.76dB for $Q$ values 3,5 and 7, respectively. Among the Super-ONN variants, the model with $Q = 3$ achieves the top performance. Most importantly, the best Super-ONN model even surpasses the 17-layer deep DnCNN by a significant margin of 1.22 dB. Qualitatively speaking, the superior denoising performance of Self-ONNs with super-neurons is once again visible in all output images shown in Figure 4. Similar comments as in the AWGN denoising results can be made here, i.e., Super-ONNs not only achieves a sharper edge and texture restoration but also recover the smooth regions better than other networks.

Table 3. Average PSNR levels of real-world denoising for DnCNN with 17 layers [ref], and denoising networks with 2 hidden layers (CNN, Self-ONN and Super-ONN) over the SIDD benchmark dataset.

| Network | Param. No (k) | Average PSNR (dB) |
| --- | --- | --- |
| CNN-64 | 297,60 | 35,167 |
| Self-ONN-3-64 | 224,83 | 35,47 |
| Self-ONN-5-64 | 374,59 | 35,52 |
| Self-ONN-7-64 | 524,35 | 35,34 |
| Super-ONN-3-64 | 224,83 | **37,28** |
| Super-ONN-5-64 | 374,59 | 37,01 |
| Super-ONN-7-64 | 524,35 | 37,1 |
| DnCNN [13] | 556,03 | 36.06 |

## 6. Discussion

### 6.1. Denoising Performance

All denoising results indicate that Super-ONNs conclusively outperform the Self-ONNs by a significant margin. Moreover, the quantitative results are supported by the qualitative performance where the restored images with Super-ONNs have a superior visual quality especially on the edges, texture and smooth regions. As reported in [11], [18], the generalization performance of Self-ONNs shows negligible improvement when increasing the value of $Q$. However, for Super-ONNs even with $Q = 3$, improving the receptive field by means of non-localized kernels provides a significant surge in performance. This implies that denoising benefits more from an enhanced receptive field as compared to sole increasing the nonlinearity level of the network. This could also be possible because of the compact networks used in the study, as there might be a saturation point for such shallow networks beyond which increasing the nonlinearity level does not necessarily benefit the denoising performance.

For real-world denoising, the performance gap by super neurons further widens over both generative and convolutional neurons. A key challenge when denoising color images is to accurately recover the color information for each pixel which can get altered even if one of the color channels in the output image is not recovered properly. As information in local neighborhoods can be corrupted severely because of noise, it becomes even more important to gather information from a larger region of the input to provide the network with the best chance of retrieving lost color information. A litmus test for this particular problem is the recovery of smooth areas in denoised images, which can appear stained if the color information is inadequate. From Figure 4, we can observe that the outputs of Super-ONNs have the least amount of color artifacts in the flat regions of the image and in some cases, recovers the color without any noticeable staining. A particular example of this is also shown in Figure 5. As depicted, the area under consideration is discolored because of the noise which causes a spot of false (purple) color. Both the CNN and Self-ONN network fail to recover the true color properly because the information from the surrounding areas is limited. In the case of Super-ONNs, we can see that the same area respects the original color better and is able to provide a superior result.

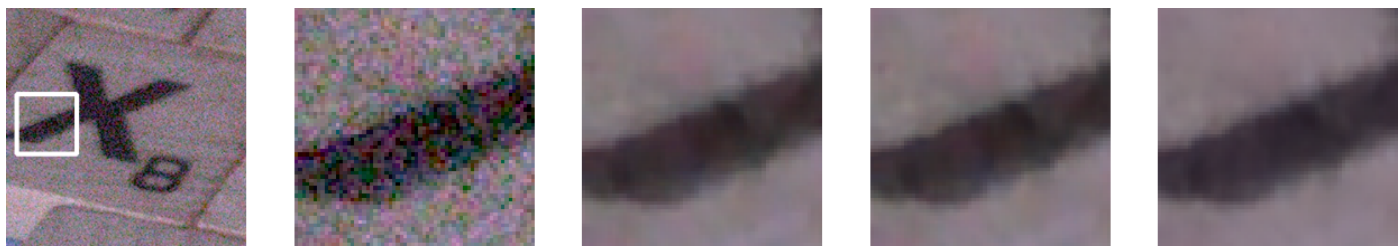

Figure 5. Example of better recovery of color information with Super-ONNs

## 6.2. Enhanced Receptive Field

Analytical calculation of receptive field size has been formulated in recent works [4] and is useful for obtaining theoretical bounds for the receptive field of a particular network configuration. However, numerical estimation of the effective receptive field size [3] using trained networks provides a more accurate understanding of the contribution of input pixels towards producing the denoised pixels in the output. To further analyze the enhanced receptive field yielded by the super neuron model, we calculate the effective receptive fields of all trained models using a simple gradient-based estimation technique as used in [3]. Specifically, a custom map with only a particular area of interest activated is backpropagated through the network. The gradient of the input of the network then provides an estimated heat map of the contribution of input pixels towards producing the network output. We plot the effective receptive field and visualize the contribution of input pixels towards denoising a particular patch of the noisy image in Figure 7. It is evident from the plots that the trained Super-ONNs enable contributions from a larger part of the input image as compared to CNNs and Self-ONNs. We can also see how this directly benefits the denoising output as the Super-ONN can recover most of the content including major edges and smooth flat regions.

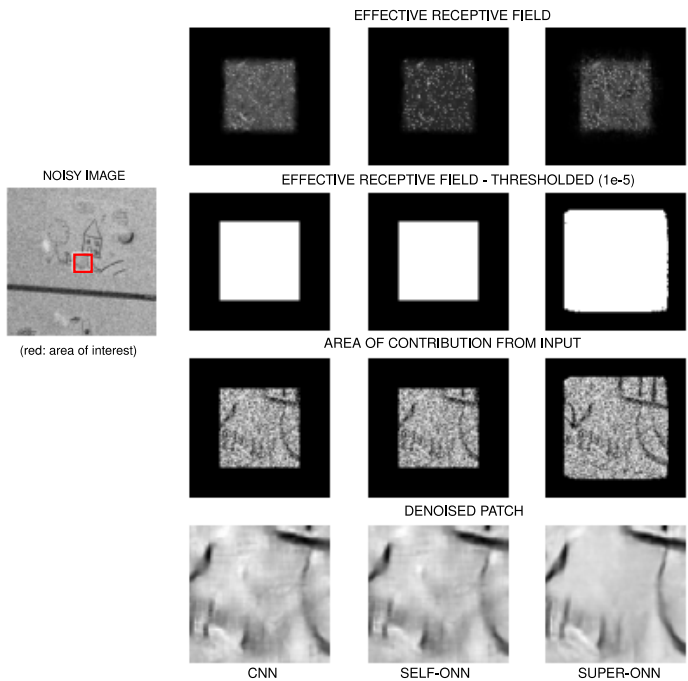

Figure 6. The effective receptive field for best CNN, Self-ONN, and Super-ONN models. The noisy image with the area of interest highlighted is shown in the first column (left). On the right, the original and thresholded effective receptive field (Row 1 and 2), masked contribution from the input image (Row 3), and the denoised patch are shown.

## 6.3. Effect of bias range on denoising

The choice of the bias range hyperparameters $(\gamma_x, \gamma_y)$, for super neurons, is critical as it directly controls the extent of non-local information integrated by the neuron. While the naïve approach towards increasing the receptive field is to simply have a large value of $(\gamma_x, \gamma_y)$, it is practically not the best approach. Specifically, shifting the kernels away from the center beyond a reasonable limit also increases the likelihood of including non-relevant information from other parts of the image. Given a training patch size $(H, W)$, it is also important to keep the ratios, $\frac{\gamma_x}{H}$ and $\frac{\gamma_y}{W}$ sufficiently small so that the kernel location does not exceed the boundaries of the image which will result in the kernel weights being optimized inappropriately. The experimental results validate this analysis as shown in Figure 7. The best results are obtained when the ratio of the maximum shift to the size of the patch is around 0.125. Even when increased slightly from this point, the denoising performance starts to decline and eventually nosedives around 0.5. In our experiments, optimizing, $\gamma_x$ and $\gamma_y$ as proposed in [12] does not have a positive impact on this performance and the trend shown in Figure 7 persists.

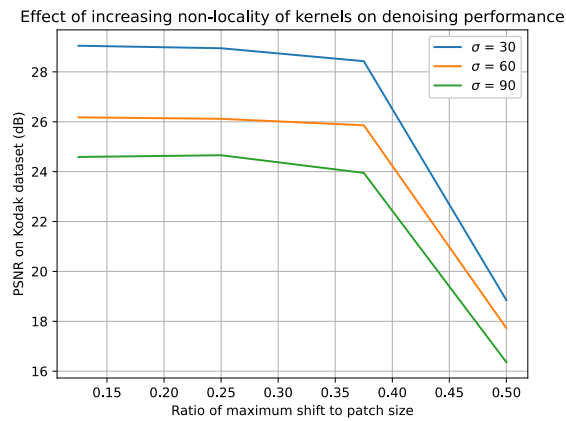

Figure 7. Effect of increasing non-locality of kernels on the denoising performance on Kodak dataset.

## 7. Conclusions

In this study, a novel image denoising methodology based on Self-ONNs empowered with super-neurons is proposed and the discriminative learning ability of super neurons for the image denoising problem was analyzed. The non-locality ability of the kernels enabled by super neurons improves the denoising performance significantly for compact networks in both synthetic and blind real-world denoising. In the case of the latter, the compact network with super neurons even significatnly outperform a recent denoiser based on a deep CNN. Furthermore, empirical estimation on trained models visually confirms the enhanced receptive field yielded by the super neuron model. Hence, this makes it the first study ever demonstrating that a deep network configuration is not obligatory for a superior denoising performance level. In other words, a compact network can still achieve a similar or better denoising performance of a deep CNN denoiser providing that the network can achieve an elegant diversity with nonlinear kernel operators and receptive field with non-localized kernels. Regarding computational efficiency, we notice that it is not straightforward to directly implement the original super neuron model on GPUs like generative neurons. A partial scheme with shared shifts that allows leveraging the generative neuron implementation was proposed as an alternative. An efficient implementation with the original super neuron model may provide even better denoising performance especially when the bias pairs for each kernel is optimized. Finally, the performance of super neurons was analyzed with respect to the kernel bias range and patch resolution. Exploring best ways to optimize the extent of non-localization for each layer will be an interesting direction for our future work.

## 8. References

[1] K. Dabov, A. Foi, V. Katkovnik, and K. Egiazarian, "Image denoising by sparse 3D transform-domain collaborative filtering," *IEEE Transactions on Image Processing*, vol. 16, no. 8, pp. 2080–2095, Jul. 2007, doi: 10.1109/TIP.2007.901238.

[2] A. Buades, B. Coll, and J. M. Morel, "A non-local algorithm for image denoising," *Proceedings - 2005 IEEE Computer Society Conference on Computer Vision and Pattern Recognition, CVPR 2005*, vol. II, pp. 60–65, 2005, doi: 10.1109/CVPR.2005.38.

[3] W. Luo, Y. Li, R. Urtasun, and R. Zemel, "Understanding the effective receptive field in deep convolutional neural networks," 2016.

[4] T. Wang, M. Sun, and K. Hu, "Dilated deep residual network for image denoising," *Proceedings - International Conference on Tools with Artificial Intelligence, ICTAI*, vol. 2017-November, pp. 1272–1279, Jun. 2018, doi: 10.1109/ICTAI.2017.00192.

[5] K. Zhang, W. Zuo, and L. Zhang, "FFDNet: Toward a fast and flexible solution for CNN-Based image denoising," *IEEE Transactions on Image Processing*, vol. 27, no. 9, pp. 4608–4622, Sep. 2018, doi: 10.1109/TIP.2018.2839891.

[6] Y. Tai, J. Yang, X. Liu, and C. Xu, "MemNet: A Persistent Memory Network for Image Restoration," in *Proceedings of the IEEE International Conference on Computer Vision*, 2017, vol. 2017-October, doi: 10.1109/ICCV.2017.486.

[7] S. Yu, B. Park, and J. Jeong, "Deep iterative down-up CNN for image denoising," in *IEEE Computer Society Conference on Computer Vision and Pattern Recognition Workshops*, 2019, vol. 2019-June, doi: 10.1109/CVPRW.2019.00262.

[8] O. Ronneberger, P. Fischer, and T. Brox, "U-net: Convolutional networks for biomedical image segmentation," in *Lecture Notes in Computer Science (including subseries Lecture Notes in Artificial Intelligence and Lecture Notes in Bioinformatics)*, 2015, vol. 9351, doi: 10.1007/978-3-319-24574-4_28.

[9] S. Kiranyaz, T. Ince, A. Iosifidis, and M. Gabbouj, "Operational neural networks," *Neural Computing and Applications 2020 32:11*, vol. 32, no. 11, pp. 6645–6668, Mar. 2020, doi: 10.1007/S00521-020-04780-3.

[10] S. Kiranyaz, J. Malik, H. Ben Abdallah, T. Ince, A. Iosifidis, and M. Gabbouj, "Self-organized Operational Neural

[11] J. Malik, S. Kiranyaz, and M. Gabbouj, "Self-organized operational neural networks for severe image restoration problems," *Neural Networks*, vol. 135, 2021, doi: 10.1016/j.neunet.2020.12.014.

[12] S. Kiranyaz, J. Malik, M. Yamac, E. Guldogan, T. Ince, and M. Gabbouj, "Super Neurons," 2020.

[13] K. Zhang, W. Zuo, Y. Chen, D. Meng, and L. Zhang, "Beyond a Gaussian denoiser: Residual learning of deep CNN for image denoising," *IEEE Transactions on Image Processing*, vol. 26, no. 7, 2017, doi: 10.1109/TIP.2017.2662206.

[14] S. Kiranyaz, J. Malik, H. Ben Abdallah, T. Ince, A. Iosifidis, and M. Gabbouj, "Exploiting heterogeneity in operational neural networks by synaptic plasticity," *Neural Computing and Applications*, vol. 33, no. 13, 2021, doi: 10.1007/s00521-020-05543-w.

[15] J. Malik, S. Kiranyaz, M. Yamac, E. Guldogan, and M. Gabbouj, "Convolutional versus Self-Organized Operational Neural Networks for Real-World Blind Image Denoising," pp. 1–25, Mar. 2021, Accessed: Aug. 16, 2021. [Online]. Available: https://arxiv.org/abs/2103.03070v2.

[16] "torch.nn.functional.grid_sample — PyTorch 1.9.0 documentation." https://pytorch.org/docs/stable/generated/torch.nn.functional.grid_sample.html#torch.nn.functional.grid_sample (accessed Aug. 24, 2021).

[17] A. Abdelhamed, S. Lin, and M. S. Brown, "A High-Quality Denoising Dataset for Smartphone Cameras," 2018, doi: 10.1109/CVPR.2018.00182.

[18] J. Malik, S. Kiranyaz, M. Yamac, and M. Gabbouj, "BM3D vs 2-Layer ONN," Mar. 2021, Accessed: Aug. 24, 2021. [Online]. Available: https://arxiv.org/abs/2103.03060v1.

[19] J. Malik, S. Kiranyaz, and M. Gabbouj, "FastONN -- Python based open-source GPU implementation for Operational Neural Networks," Jun. 2020, Accessed: Aug. 24, 2021. [Online]. Available: https://arxiv.org/abs/2006.02267v1.

[20] A. Paszke *et al.*, "PyTorch: An Imperative Style, High-Performance Deep Learning Library," *Advances in Neural Information Processing Systems*, vol. 32, 2019.

[21] M. Maggioni, V. Katkovnik, K. Egiazarian, and A. Foi, "Nonlocal transform-domain filter for volumetric data denoising and reconstruction," *IEEE Transactions on Image Processing*, vol. 22, no. 1, 2013, doi: 10.1109/TIP.2012.2210725.

Networks with Generative Neurons," *Neural Networks*, vol. 140, 2021, doi: 10.1016/j.neunet.2021.02.028.